\documentclass{article}
\usepackage{spconf,amsmath,graphicx}

\usepackage{multirow}
\usepackage{hhline}
\usepackage{multicol, blindtext}
\usepackage{array}
\usepackage{placeins}
\usepackage{float}

\usepackage{makecell}
\usepackage{array}
\usepackage{lipsum}

\ninept

\title{An Iterative Deep Learning Framework for Unsupervised Discovery of Speech Features and Linguistic Units with Applications on Spoken Term Detection}
\makeatletter
\def\name#1{\gdef\@name{#1\\}}
\makeatother
\name{{ \em Cheng-Tao Chung$^{*1}$, Cheng-Yu Tsai$^{\#2}$, Hsiang-Hung Lu$^{\#3}$, }\\
{ \em  Chia-Hsiang Liu$^{*4}$, Hung-yi Lee$^{*5}$ and Lin-shan Lee$^{\#6}$}}

\address{
Graduate Institute of Electrical Engineering, National Taiwan University$^*$ \\
Graduate Institute of Communication Engineering, National Taiwan University$^\#$}

\begin{document}

\maketitle
  \begin{abstract}
In this work we aim to discover high quality speech features and linguistic units directly from unlabeled speech data in a zero resource scenario. 
The results are evaluated using the metrics and corpora proposed in the Zero Resource Speech Challenge organized at Interspeech 2015. A Multi-layered Acoustic Tokenizer (MAT) was proposed for automatic discovery of multiple sets of acoustic tokens from the given corpus.
Each acoustic token set is specified by a set of hyperparameters that describe the model configuration. 
These sets of acoustic tokens carry different characteristics fof the given corpus and the language behind, thus can be mutually reinforced. 
The multiple sets of token labels are then used as the targets of a Multi-target Deep Neural Network (MDNN) trained on low-level acoustic features.
Bottleneck features extracted from the MDNN are then used as the feedback input to the MAT and the MDNN itself in the next iteration. 
We call this iterative deep learning framework the Multi-layered Acoustic Tokenizing Deep Neural Network (MAT-DNN), which generates both high quality speech features for the Track 1 of the Challenge and acoustic tokens for the Track 2 of the Challenge. In addition, we performed extra experiments on the same corpora on the application of query-by-example spoken term detection. The experimental results showed the iterative deep learning framework of MAT-DNN improved the detection performance due to better underlying speech features and acoustic tokens.
\end{abstract}


\section{Introduction}
%
%
In the era of big data, huge quantities of raw speech data is easy to obtain, but annotated speech data remains hard to acquire.  
%
This leads to the increased importance of zero resource applications where annotated data is not required, such as query-by-example spoken term detection. 
With the dominant paradigm of  automatic speech recognition (ASR) technologies being supervised learning \cite{hinton2012deep}, speech technologies under the zero resource scenario is a relatively less explored topic.
The goal of the Zero Resource Speech Challenge organized in Interspeech 2015 is to inspire the development of speech technologies under the extreme situation where a whole language has to be learned from scratch \cite{lee2012nonparametric,siu2014unsupervised,kamperunsupervised,levin2015segmental}. 
In this work we develop new approaches for unsupervised discovery of speech features and linguistic unit, and in the tests use the evaluation metrics and the corpora defined by the Challenge for easier comparison of the results. 
%
Track 1 of the Challenge was to construct framewise speech features representing the speech sounds that is more robust to within-speaker and across-speaker variation. 
Track 2 of the Challenge then focuses on the discovery of word linguistic units and extracting timing information for such units from the speech corpus.
In both tracks a complete set of evaluation metrics as well as a set of standard corpora were defined in order to analyze the quality of the discovered framewise speech features and linguistic units in an  different aspects without considering the backend applications. 
%
%
In addition to the metrics defined by the Challenge, a whole set of experiments on query b spoken term detection was performed to demonstrate that discovered acoustic tokens and speech features work well in a real applications .  

In this paper, we propose a completely unsupervised iterative deep learning framework for the task. A Multi-layered Acoustic Tokenizer (MAT) is used to generate multiple sets of acoustic tokens, each with a specific model configuration referred to as a layer. The different layer of the tokens carry complementary knowledge about the corpus and the language behind \cite{chung2014unsupervised}, thus can be further mutually reinforced \cite{chung2015enhancing}. The multi-layered token labels generated by the MAT are then used as the training targets of a Multi-target Deep Neural Network \cite{vu2014investigating} (MDNN) to learn the framewise bottleneck features \cite{vesely2012language} (BNFs). The BNFs are then used as feedback input to both the MAT and the MDNN in the next iteration. 
The whole framework is referred to as a Multi-layered Acoustic Tokenizing Deep Neural Network (MAT-DNN). In addition to evaluating the results with the metrics defined in the Challenge mentioned above we perform an additional set of experiments focused on query-by-example spoken term detection using the acoustic tokens discovered.  



\begin{figure*}[tbh]
\centerline{\includegraphics[width=1.0\textwidth]{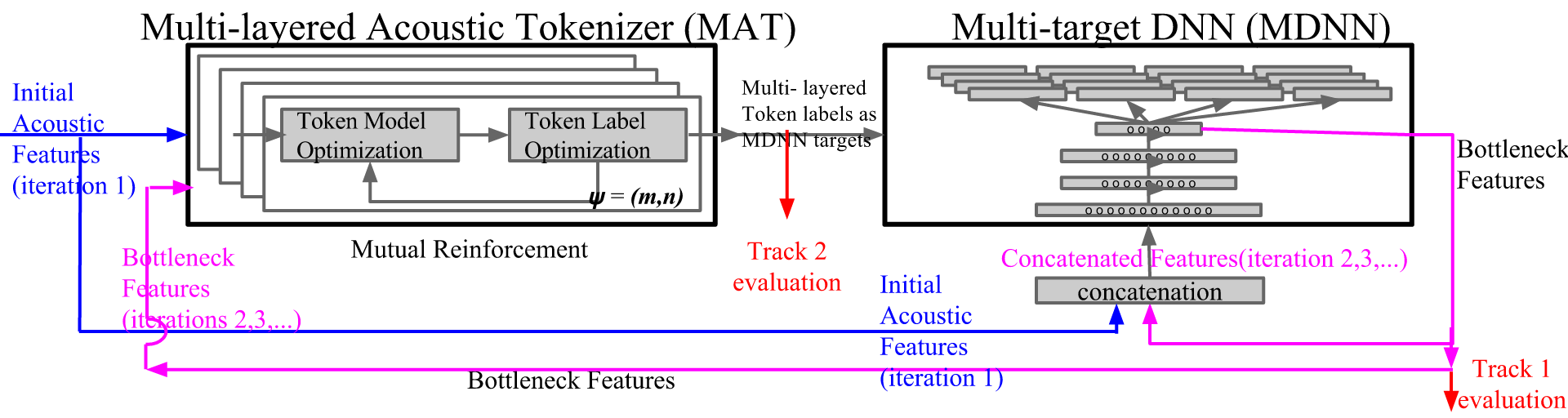}}
\caption{The proposed framework of Multi-layered Acoustic Tokenizing Deep Neural Network (MAT-DNN)}\label{fig:1}
\end{figure*}

\section{Proposed Approach}

\subsection{Overview of the proposed framework}
The framework of the approach is shown in Fig\ref{fig:1}. In the left part, the Multi-layered Acoustic Tokenizer (MAT) produces many sets of acoustic tokens using unsupervised HMMs, each describing some aspects of the given corpus. These tokens are specified by two hyperparameters $\psi=(m,n)$ describing HMM configurations which will be explained below. Each set of acoustic tokens for each configuration is obtained by iteratively optimizing the token models and the token labels on the given acoustic corpus. Multiple pairs of hyperparameters were selected producing multi-layered token labels for the given corpus to be used as the training targets of the Multi-target Deep Neural Network (MDNN) on the right part of Fig.\ref{fig:1} 
, so the knowledge carried by different token sets on different layers are fused. Bottleneck features are then extracted from this MDNN. In the first iteration, some initial acoustic features are used for both the MAT and the MDNN. This gives the first set of bottleneck features. These bottleneck features are then used as feedback to both the MAT (to replace the initial acoustic features) and the MDNN (to be concatenated with the initial acoustic features to produce tandem features) in the second iteration. Such feedback can be continued iteratively. The complete framework is referred to  as Multi-layered Acoustic Tokenizing Deep Neural Network (MAT-DNN). The output of the MDNN (bottleneck features) is evaluated with metrics of Track 1 of the Challenge, while the time intervals for the acoustic token labels at the output of the MAT are evaluated with metrics of Track 2 of the Challenge. Both the acoustic tokens and acoustic features are further examined in a query by example spoken term detection experiment in the end. 


\subsection{Multi-layered Acoustic Tokenizer(MAT)}

\subsubsection{Unsupervised Token Discovery for Each layer of MAT}
\label{sec:2-2-1}
The goal here is to obtain  multiple sets of acoustic tokens in a completely unsupervised way, each defined by the hyperparameters $\psi=(m,n)$. 
It is straightforward to discover acoustic tokens from the corpus for a chosen hyperparameter set $\psi=(m,n)$ that determines the HMM configuration (number of states per model $m$ and number of distinct models $n$)  \cite{jansen2011towards,gish2009unsupervised,siu2010improved,chung2013unsupervised,creutz2007unsupervised}.
This can be achieved by first finding an initial label set $\omega_0$ based on a set of assumed tokens for all features in the corpus $X$ as in (\ref{eq:1}) \cite{chung2013unsupervised}.
Then in each iteration $t$ the HMM parameters $\theta^\psi_{t}$ can be trained with the label set $\omega_{t-1}$ obtained in the previous iteration as in (\ref{eq:2}), and the new label set $\omega_{t}$ can be obtained by token decoding with the obtained parameters $\theta^\psi_{t}$ as in (\ref{eq:3}). 
\begin{eqnarray}
\omega_{0}&=& \mbox{initialization}(X),\phantom{\arg \max_{\substack{\theta^\psi}}}                                           \label{eq:1} \\ 
\theta^\psi_{t} &=& \arg \max_{\substack{\theta^\psi}} P(X|\theta^\psi,\omega_{t-1}),             \label{eq:2} \\
\omega_{t} &=& \arg \max_{\substack{\omega}} P(X|\theta^\psi_{t} ,\omega).                        \label{eq:3}
\end{eqnarray}
The training process can be repeated with enough number of iterations until a converged set of token HMMs is obtained. The processes (\ref{eq:2}),(\ref{eq:3}) are respectively referred to as token model optimization and token label optimization in the left part of Fig.\ref{fig:1}.

\subsubsection{Granularity Space of Multi-layered Acoustic Token Sets}\label{sec:2-2-2}

\begin{figure}[h]
\centerline{\includegraphics[width=0.4\textwidth]{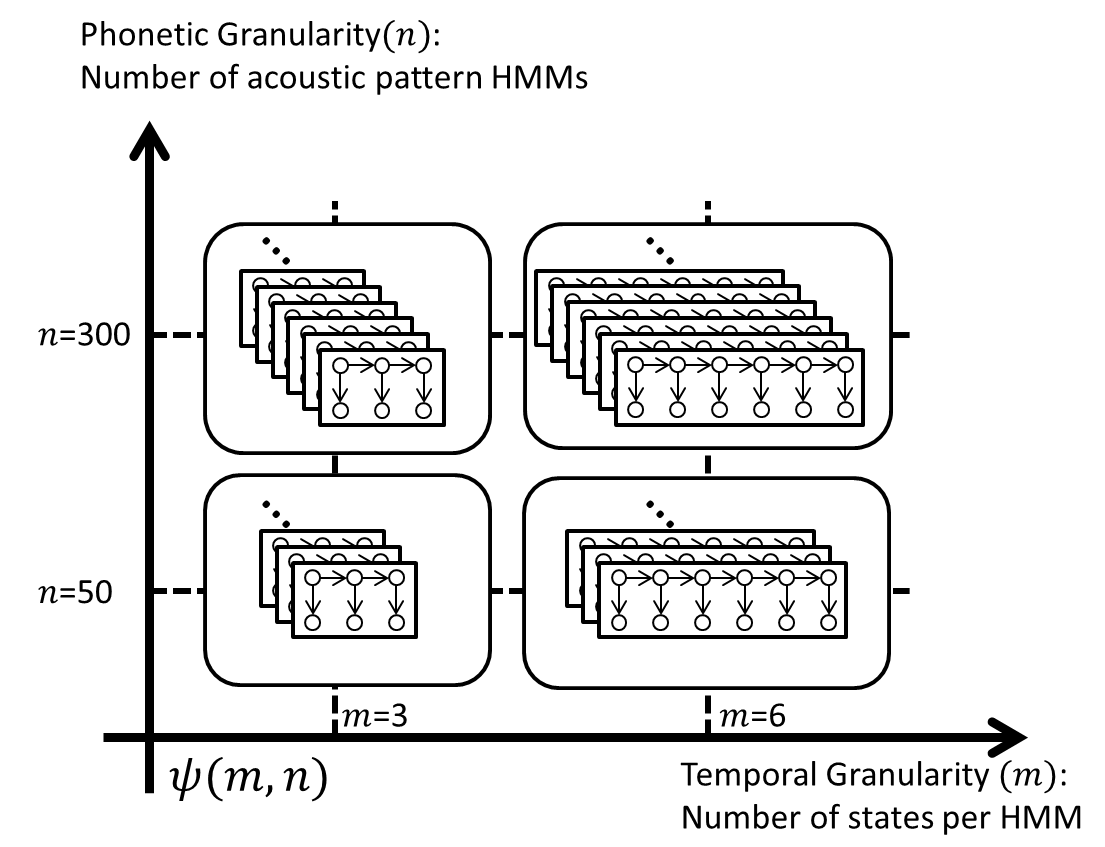}}
\caption{Model granularity space for HMM configurations}\label{fig:2dcube}
\end{figure}

The process explained above can be performed with different HMM configurations, each characterized by two hyperparameters$\psi$: the number of states $m$ in each acoustic token HMM, and the total number of distinct acoustic tokens $n$ during initialization. The token labels of a signal can be considered as a temporal segmentation, so the HMM length (or number of states in each HMM) $m$ represents the temporal granularity. The set of all distinct acoustic tokens can be considered as a segmentation of the phonetic space, so the total number $n$ of distinct acoustic tokens represents the phonetic granularity. 
This gives a two-dimensional representation in terms of temporal and phonetic granularities as in Fig.\ref{fig:2dcube}. The points in this two-dimensional space in Fig.\ref{fig:2dcube} correspond to acoustic token configurations with different model granularities, carrying complementary knowledge about the corpus and the language. 
Although the best selection of the hyperparameters in the above two-dimensional space is not known, we can simply select $M$ temporal granularities ($m$=$m_1$,$m_2$,...$m_M$) and $N$ phonetic granularities ($n$=$n_1$,$n_2$,...$n_N$), forming a two-dimensional array of $M \times N$ hyperparameter pairs in the granularity space.

\subsection{Mutual Reinforcement(MR) of Multi-layered Tokens}
Because all layers of acoustic tokens obtained in the MAT above are learned in an unsupervised fashion, they may not be very precise. But we have many layers, each for a distinct pair of hyperparameters $\psi=(m,n)$, so they can be mutually reinforced(MR). This is explained here and shown in Fig.\ref{fig:3}, including token boundary fusion and LDA-based token label re-initialization as in Fig.\ref{fig:3}(a).

\subsubsection{Token Boundary Fusion}
Fig.\ref{fig:3}(b) shows the token boundary when a part of an utterance is segmented into acoustic tokens on different layers with different hyperparameter pairs $\psi=(m,n)$. We define a boundary function $b_{m,n}(j)$ on each layer with $\psi=(m,n)$ for the possible boundary between every pair of two adjacent frames within the utterance, where $j$ is the time index for such possible boundaries. On each layer $b_{m,n}(j)$=1 if boundary $j$ is a token boundary and 0 otherwise. All these boundary functions $b_{m,n}(j)$ for all different layers are then weighted and averaged to give a joint boundary function $B(j)$. The weights consider the fact that smaller $m$ or shorter HMMs generate more boundaries, so those boundaries should weigh less. The peaks of $B(j)$ are then selected based on the second derivatives and some filtering and thresholding process. This gives the new segmentation of the utterance as shown at the bottom of Fig.\ref{fig:3}(b).

\subsubsection{LDA-based Token Label Re-initialization}
\begin{figure}[tbh]
\centerline{\includegraphics[width=0.48\textwidth]{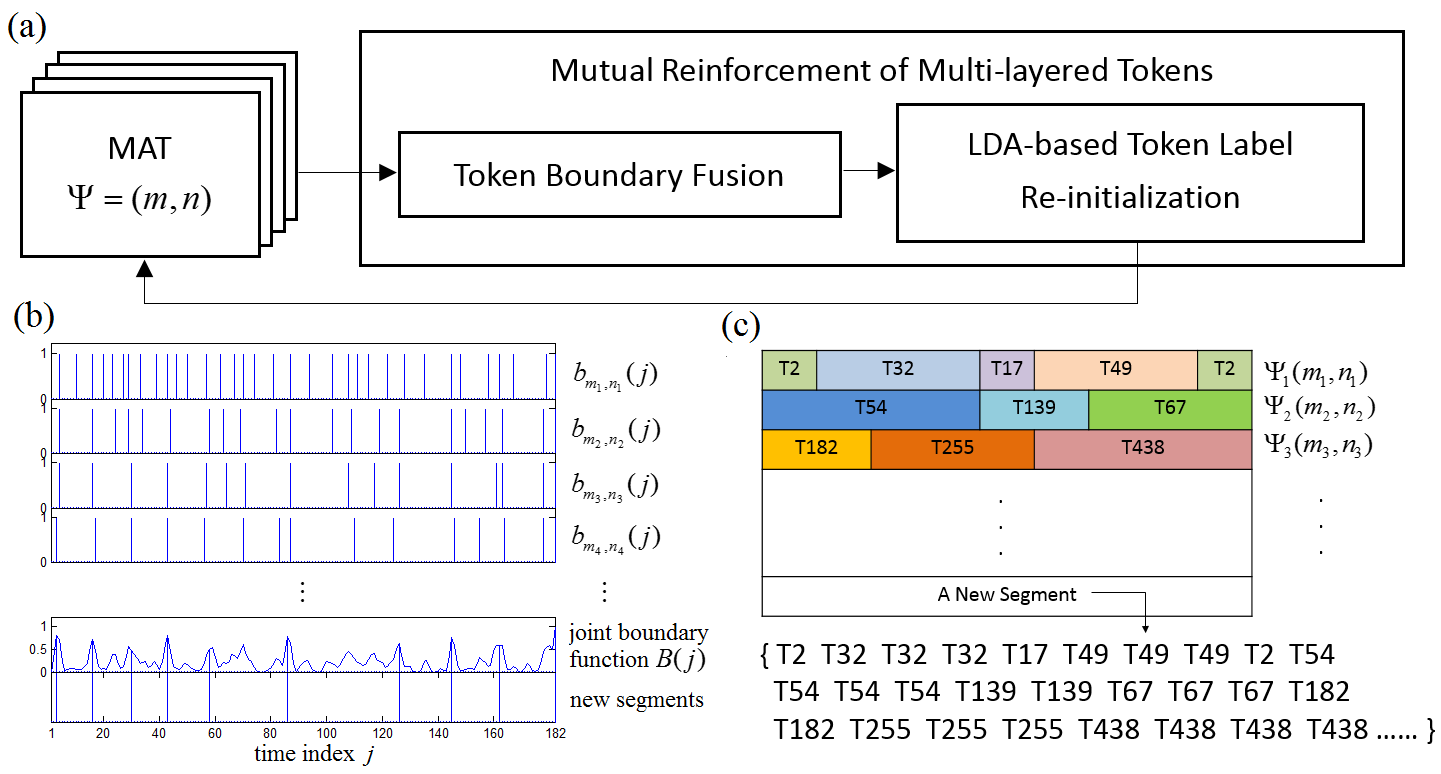}}
\caption{Mutual Reinforcement(MR) of multi-layered tokens: (a) block diagram, (b) token boundary fusion, and (c) a new segment considered as a document (bag-of-words) and a token as a word in LDA-based token label re-initialization.
}\label{fig:3}
\end{figure}

As shown in Fig.\ref{fig:3}(c), each new segment obtained above usually consists of a sequence of acoustic tokens on each layer based on the tokens defined on that layer. We now consider all the tokens on all the different layers as different words, so we have a vocabulary of $\sum\limits_{i=1}^{MN}     n_i$ words, i.e., there are $n_i$ words on the $i$-th layer and there are a total of $MN$ layers. A new segment here is thus considered as a document (bag-of-words) composed of words (tokens) collected from all different layers. Latent Dirichlet Allocation \cite{blei2003latent} (LDA) is preformed for topic modeling, and then each document (new segment) is labeled with the most probable topic. Because in LDA a topic is characterized by a word distribution, here a token distribution across different layers may also represent a certain acoustic characteristics or a certain acoustic token. By setting the number of topics in LDA as the number of distinct tokens $n$ ($n$=$n_1$,$n_2$,...$n_N$) as in subsection \ref{sec:2-2-2}, we have a new initial label set $\omega_0$ as in (\ref{eq:1}) of subsection \ref{sec:2-2-1}, in which each new segment obtained here is a new acoustic token whose ID is the topic ID obtained by LDA. This new initial label set $\omega_0$ is then used to re-train all the acoustic tokens on all layers of MAT as in (\ref{eq:1})(\ref{eq:2})(\ref{eq:3}).

\subsection{The Multi-target DNN (MDNN)}
As shown in the right part of Fig.\ref{fig:1}, token label sequence from a layer (with hyperparameters $\psi=(m,n)$) is a valid target for supervised framewise training, although obtained in an unsupervised way. In the initial work here, we simply take the token label rather than the HMM state as the training target. As shown in Fig.\ref{fig:1}, there are multi-layered token labels with different hyperparameters $\psi=(m,n)$ for each utterance offered by MAT on the left, so we jointly consider all the multi-layered token labels by learning the parameters for a single DNN with a uniformly weighted cross-entropy objective at the output layer. As a result, the bottleneck feature (BNF) extracted from this DNN automatically fuse all knowledge about the corpus and the language from the different sets of acoustic tokens.


\begin {table}[t]

\begin{center}
\begin{tabular}{|c|l|p{0.033\textwidth}|p{0.033\textwidth}|p{0.033\textwidth}|p{0.033\textwidth}|}
\hline
\multicolumn{2}{|c|}{\multirow{2}{*}{Method}} & \multicolumn{2}{c|}{English}    & \multicolumn{2}{c|}{Tsonga}     \\ \cline{3-6} 
\multicolumn{2}{|c|}{}                        & across         & within         & across         & within         \\ \hline
(1)             & Baseline                    & 28.10          & 15.60          & 33.80          & 19.10          \\ \hhline{|=|=|=|=|=|=|}
(2)             & MFCC                        & 28.63          & 15.89          & 30.77          & 16.34          \\ \hline
(3)             & DBM posterior                       & 25.96          & 15.74          & 29.15          & 16.18          \\ \hline
(4)             & BNF-1st, MR-0               & 26.84          & 15.95          & 26.48          & 15.52          \\ \hline
(5)             & BNF-1st, MR-1               & 23.88          & 14.60          & 21.97          & 13.40          \\ \hline
(6)             & BNF-1st, MR-2               & 24.46          & 14.92          & 22.14          & 13.31          \\ \hline
(7)             & BNF-2nd, MR-0               & 26.55          & 16.27          & 26.23          & 15.05          \\ \hline
(8)             & BNF-2nd, MR-1               & 24.53          & 15.13          & 23.30          & 13.88          \\ \hline
(9)             & BNF-1st, MR-1,64            & 23.15 & 14.53  & 21.91 & 13.28 \\ \hline 
(10)             & BNF-1st, MR-1,128           & 22.98 & 14.49 & 21.87 & 13.28 \\ \hline
(11)             & BNF-1st, MR-1,256            & \textbf{21.92}  & \textbf{13.95} & \textbf{21.42} & \textbf{12.84} \\ \hline
(12)            & BNF-2nd, MR-1*            & 24.13           & 15.24          & 
23.05          & 14.03          \\ \hhline{|=|=|=|=|=|=|}
(13)            & Topline                     & 16.00          & 12.10          & 04.50          & 03.50          \\ \hline
\end{tabular}
\end{center}

\caption {Speech feature quality in the metrics of Track 1 of the Challenge. The best figure for each metric is shown in bold. \label{tab:1}}
\end{table}

\subsection{The Iterative Learning Framework for MAT-DNN}
Once the BNFs are extracted from the MDNN in iteration 1, they can be taken as the input to the MAT on the left of Fig.\ref{fig:1} replacing the initial acoustic features. The MAT then generates updated sets of multi-layered token labels and these updated sets of multi-layered token labels can be used as the updated training targets of the MDNN. The input features of the MDNN can also be updated by concatenating the initial acoustic features with the newly extracted BNFs as the concatenated features, and this process can be repeated.
The concatenated feature used as the input of the MDNN can be further augmented by concatenating unsupervised features obtained with other approaches such as the Deep Boltzmann Machine \cite{salakhutdinov2009deep} (DBM) posteriorgrams, Long-Short Term Memory Recurrent Neural Network \cite{hochreiter1997long} (LSTM-RNN) autoencoder bottleneck features, and i-vectors \cite{kanagasundaram2011vector} trained on MFCC.
Although different from the conventional recurrent neural network (RNN) in which the recurrent structure is included in back propagation training, the concatenation of the bottleneck features with other features in the next iteration in MDNN is a kind of recurrent structure.

\label{sec:2-5}

\begin{table*}[h!t]
\centering
\tabcolsep=0.11cm
\begin{tabular}{!{\vrule width 1pt}c|c|l!{\vrule width 1pt}cc!{\vrule width 1pt}ccc!{\vrule width 1pt}ccc!{\vrule width 1pt}ccc!{\vrule width 1pt}ccc!{\vrule width 1pt}ccc!{\vrule width 1pt}}
\Xhline{3\arrayrulewidth}
\multicolumn{3}{!{\vrule width 1pt}c!{\vrule width 1pt}}{\multirow{2}{*}{(\%)}}                                                               & \multirow{2}{*}{NED} & \multirow{2}{*}{Cov.} & \multicolumn{3}{c!{\vrule width 1pt}}{Matching}      & \multicolumn{3}{c!{\vrule width 1pt}}{Grouping} & \multicolumn{3}{c!{\vrule width 1pt}}{Type}          & \multicolumn{3}{c!{\vrule width 1pt}}{Token}                   & \multicolumn{3}{c!{\vrule width 1pt}}{Boundary}        \\ \cline{6-20} 
\multicolumn{3}{!{\vrule width 1pt}c!{\vrule width 1pt}}{}                                                                                    &                      &                       & P    & R            & F            & P        & R        & F       & P   & R             & F            & P            & R             & F             & P    & R             & F             \\ \Xhline{3\arrayrulewidth}
\multirow{2}{*}{Eng.} & \multicolumn{2}{c!{\vrule width 1pt}}{JHU}                                                          & 21.9                 & 16.3                  & 39.4 & 1.6          & 3.1          & 21.4     & 84.6     & 33.3    & 6.2 & 1.9           & 2.9          & 5.5          & 0.4           & 0.8           & 44.1 & 4.7           & 8.6           \\ \cline{2-20} 
                      & (A) & \begin{tabular}[c]{@{}l@{}}(4) TOK-1st, MR-0\\ $\psi=(7,50)$\end{tabular}   & 87.5                 & 100                   & 1.4  & 0.5          & 0.8          & 3.6      & 18.7     & 6       & 4.2 & \textbf{11.9} & \textbf{6.2} & \textbf{8.3} & \textbf{15.7} & \textbf{10.9} & 35.2 & \textbf{84.6} & \textbf{49.8} \\ \Xhline{3\arrayrulewidth}
\multirow{3}{*}{Tso.} & \multicolumn{2}{c!{\vrule width 1pt}}{JHU}                                                          & 12                   & 16.2                  & 69.1 & 0.3          & 0.5          & 52.1     & 77.4     & 62.2    & 3.2 & 1.4           & 2            & 2.6          & 0.5           & 0.8           & 22.3 & 5.6           & 8.9           \\ \cline{2-20} 
                      & (B) & \begin{tabular}[c]{@{}l@{}}(8) TOK-2nd, MR-1\\ $\psi=(9,50)$\end{tabular}   & 69.1                 & 95                    & 5.9  & \textbf{0.5} & \textbf{0.9} & 10.7     & 26.8     & 15.3    & 1.5 & \textbf{3.9}           & \textbf{2.2}          & 2.3          & \textbf{6.6}           & \textbf{3.4}           & 17.1 & \textbf{59.1}          & \textbf{26.6}          \\ \cline{2-20} 
                      & (C) & \begin{tabular}[c]{@{}l@{}}(5) TOK-1st, MR-1\\ $\psi=(13,300)$\end{tabular} & 60.2                 & 96.1                  & 9.7  & \textbf{0.4}          & \textbf{0.8}          & 13.5     & 12.7     & 13.1    & 1.8 & \textbf{4.7}  & \textbf{2.5} & \textbf{3.9} & \textbf{9.1}  & \textbf{5.4}  & 21.2 & \textbf{62.1} & \textbf{31.6} \\ \Xhline{3\arrayrulewidth}
\end{tabular}
\caption{Comparison of three typical example token sets selected out of all shown in Fig.\ref{fig:t1} with the JHU baseline offered by the Challenge. Those better than JHU baseline are in bold.}
\label{tab:2}
\end{table*}

\subsection{Spoken Term Detection}
\label{sec:2-6}
Let \{$p_r, r=1,2,3,..,n$\} denote the $n$ acoustic tokens in the set of $\psi$=$(m,n)$. We first construct a distance matrix $S$ of size $n \times n$ off-line for every token set $\psi$=$(m,n)$, for which the element $S(i,j)$ is the distance between any two token HMMs $p_i$ and $p_j$ in the set.
\begin{equation}
S(i, j) =\mbox{KL}(i, j). \label{eq:soft}
\end{equation}
The KL-divergence $\mbox{KL}(i,j)$ between two token HMMs in (\ref{eq:soft}) is defined as the symmetric KL-divergence between the states based on the variational approximation \cite{hershey2007approximating} summed over the states. 

In the on-line phase, we perform the following for each entered spoken query $q$ and each document (utterance) $d$ in the archive for each token set $\psi$=$(m,n)$. Assume for a given pattern set a document $d$ is decoded into a sequence of $D$ acoustic patterns with indices $(d_1, d_2, ..., d_D)$ and the query $q$ into a sequence of $Q$ patterns with indices $(q_1, ..., q_Q)$. 
We thus construct a matching matrix $W$ of size $D \times Q$ for every document-query pair, in which each entry $(i,j)$ is the distance between acoustic tokens with indices $d_i$ and $q_j$ as in (\ref{eq:topk}), where $S(i,j)$ is defined in (\ref{eq:soft}),
\begin{equation}
W(i, j)  = S(d_i, q_j).  \label{eq:topk}
\end{equation}
We perform token-based DTW on this matching matrix $W$ by summing the distance between token pairs along the optimal path and return the minimal distance as the distance between document $d$ and query $q$.

\section{Experimental Setup}

The MAT-DNN presented above allows flexible configurations, but here we train the MAT-DNN in the following manner. We set $m$=3, 5, 7, 9 states per token HMM and $n$=50, 100, 300, 500 distinct tokens in the MAT, which gave a total of 16 layers($m$=11,13 were added in some tests as mentioned below).

In the first iteration, we used the 39 dimension Mel-frequency Cepstral Coefficients (MFCC) with energy, delta and double delta as the initial acoustic features for the input to both the MAT and the MDNN.  We concatenated the MFCC with a window of 4 frames before and after (39x9 dimensions), and an i-vector (400 dimensions) trained on the MFCC of each evaluation interval used as the input of the MDNN. The topology of the MDNN is set to be 751(input)-256(hidden)-256(hidden)-39(bottleneck)-(target) with 3 hidden layers. 
and we kept the dimensionality of these features to be 39 for a fair comparison. For the Deep Boltzmann Machine(DBM), we used the 39-dimension MFCC with a window of 5 frames before and after as the input. The configuration we used for the DBM is 429(visible)-256(hidden)-256(hidden)-39(hidden). We also extracted another set of LSTM-RNN autoencoder bottleneck features but found the performance was slightly worse than MFCC.

In the second iteration, we concatenated the original MFCC, the BNF extracted from the first iteration, the DBM posteriorgrams, and the i-vector forming a (39x9+39x9+39x9+400=1453) dimension input to the MDNN. We used the updated token labels as the target and extracted the BNF as the features. 

The MAT was trained using the zrst \cite{chung2014zero}, a python wrapper for the HTK toolkit \cite{young1997htk} and srilm \cite{stolcke2002srilm} that we developed for training unsupervised HMMs with varying model granularity. The LDA model we used in the Mutual Reinforcement was trained by MALLET \cite{mccallum2002mallet}. The MFCC were extracted using the HTK toolkit \cite{young1997htk}. The i-vectors were extracted using Kaldi \cite{Povey_ASRU2011}. The DBM posteriorgram was extracted using libdnn \cite{chou2014libdnn}. The MDNN was trained using  Caffe \cite{jia2014caffe}. The two corpora used in the Challenge were used here: the Buckeye corpus \cite{pitt2007buckeye} in English and the NCHLT Xitsonga Speech corpus in Tsonga.

\section{Experimental Results}
\subsection{Feature Quality in Metrics of Track 1}

The evaluation was based on the ABX discriminability test \cite{schatz2013evaluating} including across-speaker and within-speaker tests. 
The warping distance obtained by performing Dynamic Time Warping on feature sequences of predefined phone pairs was used as the distance metric for the ABX discriminability test.
The results in error percentage(the lower the better) are listed in Table \ref{tab:1}.

\begin{figure*}[th!]
\centerline{\includegraphics[width=\textwidth]{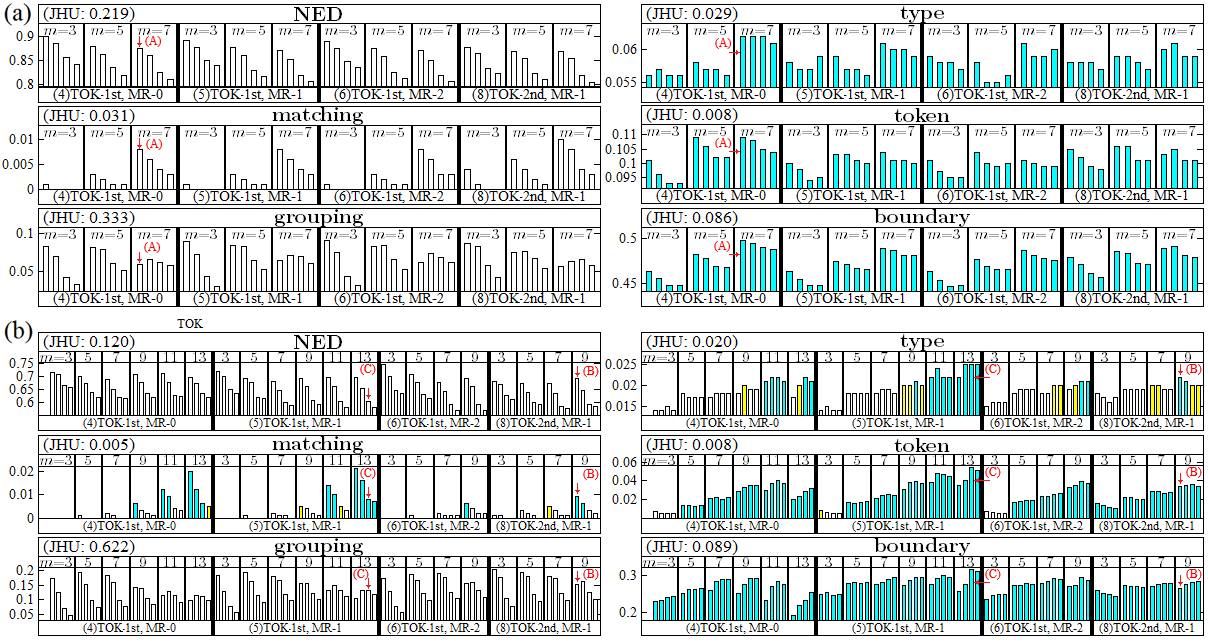}}
\caption{Quality of discovered units in metrics of Track 2 for (a) English and (b) Tsonga. Each subgraph is for an evaluation measure, including four sections from left to right for four cases of token sets used to train the bottleneck features listed in four rows of Table \ref{tab:1} as marked at the bottom. The four bars in each group for a value of $m$ are for $n$=50, 100, 300, 500 from left to right (not shown in the figure) and $\psi=(m,n)$ are parameters for the token sets. Blue, yellow and white bars correspond to better, equal to or worse as compared to the JHU baseline offered by the Challenge listed at the upper left corner of each subgraph. The coverage is not shown because it is almost 100\% in all cases.}
\label{fig:t1}
\end{figure*}

Rows (1) and (13) are the official baseline MFCC features and the official topline supervised phone posteriorgrams provided by the Challenge respectively. Row (2) is our baseline of the MFCC features, the initial acoustic features used to train all systems in this work. Row (3) is for the DBM posteriorgrams extracted from the MFCC of row (2), serving as a strong unsupervised baseline. The results in rows (4), (5) and (6) are the performance of the bottleneck features(BNF-1st) extracted in the first iteration of the MAT-DNN without applying Mutual Reinforcement (MR) (4), applying MR once(MR-1) (5), and twice(MR-2) (6) respectively.
Rows (7) and (8) are the same as rows (4), (5) except the bottleneck features were extracted in the second iteration of the MAT-DNN (BNF-2nd) and
the MAT of the MAT-DNN was trained using the BNF of row(5).
Row (9),(10) and (11) is similar to row (5), except we use a wider bottleneck layer with 64, 128, 256 dimensions instead of 39.
Row (12) is similar to row (8), except only the MFCC and i-vectors were concatenated as input without other features.


All the features from row (2) to (10) except for (9) are confined to 39 dimensions for fair comparison. We observe that as a stand-alone feature extractor without any iterations, the MAT-DNN in row (5) outperformed the DBM baseline in (3). The effect of Mutual Reinforcement can be seen in the improvement from rows (4) to (5)(6) and (7) to (8). We observe that a single iteration of Mutual Reinforcement is enough to bring huge improvement. 
The effect of iterations in the MAT-DNN can be seen by comparing rows (2), (5), (8), respectively corresponding to 0, 1, and 2 iterations. 
Although the performance improvement from row (2) to row (5) is notable, it dropped in the second iteration in (8). 
To investigate reasons of this performance drop, we widened the bottleneck feature to 256 dimensions in row (9) and observed a dramatic improvement. It is possible that we have not yet explored the full potential of the MAT-DNN as comparison between algorithms was the original goal when we designed the experiments. For a better tuned set of parameters, improvement in following iterations is to be expected on Track 1. Nonetheless, the benefit of the second iteration is better observed in Track 2.

\subsection{Quality of the Discovered Units in Metrics of Track 2}
Track 2 of the Challenge defined a total of 7 evaluation metrics for 3 tasks \cite{ludusan2014bridging} describing different aspects of the quality of the linguistic unit discovered from the corpora: coverage, Normalized Edit Distance(NED) and matching F-score for the matching task; grouping F-score and type F-score for clustering task; token F-score, boundary F-score for the parsing task. 
Except for coverage and NED whose values are indicators of system characteristic rather than a system performance, the higher the value the better for the other five metrics.
Except for coverage, the other six scores are shown in the six subfigures in Fig.\ref{fig:t1}(a) and (b). We omit coverage because it is always 100\% in all cases.
In each subfigure, the results for four cases are shown in four sections from left to right, corresponding to the four sets of tokens obtained in MAT after the first and second iterations of MAT-DNN (TOK-1st or TOK-2nd) with MR performed or not (MR-0,1,2). 
The corresponding bottleneck features for them are in front of those listed in rows (4), (5), (6) and (8) of Table \ref{tab:1}, as marked at the bottom of each section.
For each of these section, the three or six groups of bars correspond to different values of $m$ ($m$=3, 5, 7 or $m$=3, 5, 7, 9, 11, 13), while in each group the four bars correspond to the four values of $n$ ($n$=50, 100, 300, 500 from left to right), where $\psi=(m,n)$ are the parameters for the token sets. 
The bars in blue and yellow are those better or equal to the JHU baseline offered by the Challenge, while those in white are worse. Only the results jointly considering both within and across talker conditions are shown.

From Fig.\ref{fig:t1}(a) for English, it can be seen that the proposed token sets perform well in type, token and boundary scores, although much worse in matching and grouping scores. We see in many cases the benefits brought by MR (e.g. MR-2 in (6) vs MR-1 in (5) in type of Fig.\ref{fig:t1}(a)) and the second iteration (e.g. TOK-2nd in (8) vs TOK-1st in (5) in boundary of Fig.\ref{fig:t1}(a)), especially for small values of $m$. In many groups for a given $m$, smaller values of $n$ seemed better, probably because $n$=50 is close to the total number of phonemes in the language. Also, a general trend is that larger values of $m$ were better, probably because HMMs with more states were better in modeling the relatively long units; this may directly lead to the higher type, token and boundary scores.

Similar observations can be made for Tsonga in Fig.\ref{fig:t1}(b), and the overall performance seemed to be even better as the proposed token sets performed well even in matching scores. The improvements brought by MR (e.g. MR-1 vs MR-0), the bottleneck features (compared to JHU baseline) and the second iteration (TOK-2nd vs TOK-1st) are better observed here, which gave the best cases for all the five main scores. This is probably due to the fact that more sets of tokens were available for MR and MAT-DNN  on Tsonga than English. We can conclude from this observation that more token sets introduces more robustness and that leads to better token sets for the next iteration.  
When $m$ goes to 13, we see that for MR-0 in the left section of Fig.\ref{fig:t1}(b)) almost all metrics degraded except for matching scores, but with MR-1, MR-2 almost all the scores consistently increases (except for NED) when $m$ became larger. This suggests that MR can prevent degradation from happening while detecting relatively long units.

We selected three typical example token sets (A)(B)(C) out of the many proposed here, and compared them with the JHU baseline \cite{jansen2011efficient} in Table \ref{tab:2} including Precision (P), Recall (R) and F-scores (F). These three example sets are also marked in Fig.\ref{fig:t1}. In Table \ref{tab:2} those better than JHU baseline are in bold. The much higher NED and coverage scores suggest that the proposed approach is a highly permissive matching algorithm. The much higher parsing scores (type, token and boundary scores), especially the Recall and F-scores, imply the proposed approach is more successful in discovering word-like units. However, the matching and grouping scores were much worse probably because the discovered tokens covered almost the whole corpus, including short pauses or silence, and therefore many tokens were actually noises. Another possible reason might be that the values of $n$ used were much smaller than the size of the real word vocabulary, making the same token label used for signal segments of varying characteristics and this degenerated the grouping qualities.  

\subsection{Unsupervised Spoken Term Detection}
Although the discovered speech features (BNFs) and linguistic units (tokens) were evaluated to be of high quality in Tables \ref{tab:1},\ref{tab:2} and Fig. \ref{fig:t1} in various aspects in terms of the metrics defined in the Challenge, in this paper we wish to investigate if the proposed MAT-DNN is good for a real application, i.e. spoken term detection. 
Separate query by example spoken term detection experiments were conducted on the two corpora, English (Eng) and Tsonga (Xit). 
For English/Tsonga, spoken instances of 5/10 query words randomly selected from the data set were used as the spoken query to search for other instances in the spoken archive. 
%
%
Both the selected queries and the corpora were first labeled as sequences of the multi-layered tokens. 
The distance between the document token sequences and query token sequence is evaluated by the token DTW distance as defined in section \ref{sec:2-6}. 
A total of 5 collections of multi-layered token sets were tested here, which are  (TOK-1st, MR-0), (TOK-1st, MR-1), (TOK-1st, MR-2), (TOK-2nd, MR-0), (TOK-2nd, MR-1). 
For English, each collection consists of 3$\times$4 sets of acoustic tokens with granularity $m$ = 3, 5, 7 and $n$ = 50, 100, 300, 500, so we obtained 12 scores for every query-document pair on every collection. 
For Tsonga, $m$ = 3, 5, 7, 9 and $n$ = 50, 100, 300, 500, thus we had 16 scores for every query-document pair. 
We averaged the 12, or 16 distances in every collection and obtained the results. 
Mean Average Precision(MAP), the higher the better, was used as our evaluation metric, and dynamic time warping on the feature sequences was taken as the baseline.

The results for the 5 collections of tokens are in row (a) to (e) in \ref{tab:x}. 
The benefit of the iterative framework of Mutual Reinforcement (MR) can be observed by comparing rows (a) to (b), (b) to (c) and (b) to (d) (MR-0 vs MR-1, MR-1 vs MR-2). 
The benefit of the iterative framework of the MAT-DNN can be observed by comparing row (a) to (d) and (b) to (e) (TOK-1st vs TOK-2nd). 
We then averaged all token DTW distances in (a) to (e) in row(h), and obtained better results, showing that the information obtained in each collection is complimentary to each other as well. 
We then compared these results with two cases of DTW performed on frame-level features: 39-dim MFCC in row (f) and bottleneck features (BNF-1st, MR-1) in row (g). 
By comparing rows (g) to (f), we observe that the features obtained by MAT-DNN performed significantly better than the MFCC from which they were derived. 
We further fused the information from both the feature based DTW and token based DTW by averaging all scores in rows  (a) to (g) in row (i), producing even better results indicating frame-level and token-level information are complementary.

\begin{table}[t]
\begin{center}\begin{tabular}{|c|l|c|l|r|}
\hline
\multirow{2}{*}{method}      & \multicolumn{2}{c|}{\multirow{2}{*}{index}} & \multicolumn{2}{c|}{MAP(\%)}                        \\ \cline{4-5} 
                             & \multicolumn{2}{c|}{}                       & \multicolumn{1}{c|}{Eng} & \multicolumn{1}{c|}{Xit} \\ \hline
\multirow{5}{*}{token DTW}   & (a)             & TOK-1st, MR-0             & 12.49                    & 19.00                    \\ \cline{2-5} 
                             & (b)             & TOK-1st, MR-1             & 13.98                    & 21.27                    \\ \cline{2-5} 
                             & (c)             & TOK-1st, MR-2             & 13.42                    & 24.17                    \\ \cline{2-5} 
                             & (d)             & TOK-2nd, MR-0             & 10.37                    & 25.58                    \\ \cline{2-5} 
                             & (e)             & TOK-2nd, MR-1             & 14.51                    & 25.44                    \\ \hline
\multirow{2}{*}{feature DTW} & (f)             & MFCC                      & 11.08                    & 8.96                    \\ \cline{2-5} 
                             & (g)             & BNF-1st, MR-1                   & 13.39                    & 28.71                    \\ \hline
\multirow{2}{*}{fusion}     & (h)             & (a)-(e)                   & 15.28                    & 26.17                    \\ \cline{2-5} 
                             & (i)             & (a)-(g)                   & 18.01                    & 26.33                    \\ \hline
\end{tabular}
\end{center}
\caption {Overall spoken term detection performance in mean average precision.\label{tab:x}}
\end{table}

\section{Conclusion}
In this paper we propose an iterative deep learning framework, MAT-DNN, to discover high quality features and multi-layer acoustic token sets on a completely unsupervised way. These features and tokens are evaluated by the metrics and corpora defined in the Zero Resource Speech Challenge in Interspeech 2015. We fuse the information obtained from different token sets in the spoken term detection experiments and obtain good initial results. We hope that these results serve as good references for future investigations.

\section{Acknowledgment}
We would like to thank Yuan-ming Liou, Yen-Chen Wu, and Yen-Ju Lu for providing the DBM posteriorgrams and i-vectors used in the MAT-DNN of this work.

\bibliographystyle{IEEEbib}
\bibliography{mycap}

\end{document}